\documentclass{article}

    \usepackage[preprint]{neurips_2019}

\usepackage[utf8]{inputenc} 
\usepackage[T1]{fontenc}    
\usepackage{hyperref}       
\usepackage{url}            
\usepackage{booktabs}       
\usepackage{amsfonts}       
\usepackage{nicefrac}       
\usepackage{microtype}      

\usepackage{amsthm}
\usepackage{amsmath}
\usepackage{comment}
\usepackage{natbib}  
\usepackage{graphicx}
\usepackage{subcaption} 

\title{Formatting Instructions For NeurIPS 2019}

\author{%
  Kyle Luther \\
  Department of Physics\\
  Princeton University\\
  \texttt{kluther@princeton.edu} \\
  \And
  H. Sebastian Seung \\
  Department of Computer Science and Neuroscience \\
  Princeton University \\
  \texttt{sseung@princeton.edu} \\
}

\begin{document}

\title{ Sample Variance Decay in Randomly Initialized ReLU Networks }
\maketitle

\begin{abstract}
    Before training a neural net, a classic rule of thumb is to randomly initialize the weights so the variance of activations is preserved across layers. This is traditionally interpreted using the total variance due to randomness in both weights \emph{and} samples. Alternatively, one can interpret the rule of thumb as preservation of the variance over samples for a fixed network. The two interpretations differ little for a shallow net, but the difference is shown to grow with depth for a deep ReLU net by decomposing the total variance into the network-averaged sum of the sample variance and square of the sample mean. We demonstrate that even when the total variance is preserved, the sample variance decays in the later layers through an analytical calculation in the limit of infinite network width, and numerical simulations for finite width. We show that Batch Normalization eliminates this decay and provide empirical evidence that preserving the sample variance instead of only the total variance at initialization time can have an impact on the training dynamics of a deep network.
    
\end{abstract}

\section{ Introduction }
The procedure used to initialize the weights and biases of a neural network has a large impact on network training dynamics, and in some cases determines if the network will train at all \citep{kaiming_init}. One rule of thumb, dating back to the 90's, is to randomly initialize the weights so that the preactivations have a fixed variance in all layers of the network. In a fully connected network with $\tanh$ nonlinearity and fixed layer width $n$, this can be achieved by drawing weights from a Gaussian distribution with zero mean and standard deviation $ \sqrt{1/n} $ \citep{efficient_backprop}. The rule of thumb was extended to networks with ReLU nonlinearity by \citet{kaiming_init}, who increased the standard deviation to $ \sqrt{2/n} $ to preserve variance in the presence of ReLU. We will refer to this prescription as ``Kaiming initialization.''

In the preceding work, the rule of thumb was interpreted using the total variance of the preactivations due to the randomness in both networks (weights) and samples. Intuitively, it might be more relevant to fix the network, and focus only on the fluctuations of the preactivation over the distribution of samples. In this alternative interpretation of the rule of thumb, a network should be initialized to preserve the sample mean and variance across layers.\footnote{Note that the total mean of the preactivation is automatically preserved, because it vanishes if the weights are initialized from a distribution with zero mean.} This alternative interpretation is implicit in the data-dependent initialization scheme of \citet{data_dependent_init}.

The main goal of this paper is to demonstrate that the two interpretations of the rule of thumb can differ greatly for deep ReLU nets, although there is little difference for shallow nets. The simple but key theoretical insight is to decompose total variance into the network-averaged sum of sample variance and square of sample mean. In the limit of infinite network width, it turns out that both terms in the sum can be calculated using the technique introduced by \citet{exponential_expressivity} for studying the similarity of two activation vectors corresponding to two different samples. The propagation of this similarity through the layers of the net determines the propagation of the sample mean and variance. 

The formalism is applied to wide fully connected ReLU nets using Kaiming initialization. We show that the sample variance vanishes with depth, and the fluctuations in the sample mean across networks dominate the total variance. In other words, Kaiming initialization causes all input samples to be mapped to almost the same preactivation vector, up to small fluctuations around the sample mean. For most neurons in a deep layer, the preactivation has small fluctuations about a large positive mean or a large negative mean. For almost all samples, these neurons are either operating as if they were linear, or do not exist. We show that inserting Batch Normalization \citep{batchnorm} into a wide ReLU net changes this picture by explicitly forcing every preactivation to have 0 mean and fixed variance.

The preceding theoretical analysis suggests that the two interpretations of the rule of thumb can lead to very different initializations in deep ReLU nets with Kaiming initialization falling into the total variance-preserving camp and Batch Normalized networks falling into the sample-variance preserving camp. Whether this difference is relevant for training is unclear, and the theory as presented is not strictly applicable to finite width nets, convolutional nets or more complex architectures. Therefore we conducted experiments in which commonly used convolutional nets were trained on object recognition and image segmentation tasks. We implemented both interpretations of the rule of thumb via data-dependent initializations and additionally compared to Batch Normalized networks. In one case, we scaled the weights in each successive layer to preserve total variance. In the other case, we adjusted the biases in each successive layer so that the sample mean of the preactivation vanished, and scaled the weights so that the sample variance remains constant.\footnote{This is essentially the same as Algorithm 1 of \citet{data_dependent_init}}. We found that sample mean-variance preservation led to faster training than total variance preservation.


\section{ Background and Definitions }
For the next two sections, we assume a fully connected, feedforward network of depth $ L $ using rectified linear (ReLU) nonlinearity. This is defined for layers $ l=1,2,...,L $ by the following: 
\begin{equation}
    u^l_i = \sum_{j=1}^{n^{l-1}} W^l_{ij} x^{l-1}_j + b^l_i \;\;\;\;\;\; 
    x^l_i = f(u^l_i) \;\;\;\;\;\;
    f(x) = 
    \begin{cases}
        x,& \text{if } x\geq 0\\
        0,& \text{if } x < 0
    \end{cases}
\label{eqn:network}
\end{equation}
where $ u^l_i $ is the i'th pre-activation in layer $ l $ and $ x^l_i $ the i'th activation in layer $ l $. $ x^0_i $ are the network inputs. $ n^l $ is the width of layer $ l $. These equations are defined for all input vectors $ t=1,2,...T $ whose index has been omitted from $ u $ and $ x $ for notational clarity.

We assume networks are initialized with Kaiming initialization \citep{kaiming_init}:
\begin{equation}
    W^l_{ij} \stackrel{iid}\sim \mathcal{N}(0,\frac{2}{n^{l-1}}) \;\;\;\;\;\;\;\;\;\;\;\; b^l_i = 0  
\label{eqn:kaiming_init}
\end{equation}

\subsection{ Total Mean and Variance }

The behavior of this initialization scheme was originally demonstrated by analyzing the total mean and variance of pre-activations:
\begin{equation}
    \text{total mean: } \mu^l_i = \langle \langle u^l_i \rangle_w \rangle_t \;\;\;\;\;\;\;\;\;\;\;\ 
    \text{total variance: } (\sigma^l_i)^2 = \langle \langle (u^l_i)^2 \rangle_w \rangle_t - \langle \langle u^l_i \rangle_w \rangle_t^2 
\label{eqn:total_stats}
\end{equation}
The notation $ \langle \cdot \rangle_{w} $ denotes an average over networks and $ \langle \cdot \rangle_{t} $ an average over samples. Therefore the total mean is the mean of a pre-activation over random initializations and samples. The total variance is the variance of a pre-activation over random initializations and samples.

The expectation over $W$ implies that $\mu^l_i$ and $\sigma^l_i$ are the same for all pre-activations $i$ within a layer. For this reason we often just write $\mu^l,\sigma^l$ and refer the mean and variance of a \emph{layer} rather than a pre-activation. It was shown in \citet{kaiming_init} that for ReLU networks initialized using Equation \ref{eqn:kaiming_init}, the total mean is zero and total variance is the same for \emph{all} pre-activations, regardless of the sample distribution. For normalized inputs, i.e. $\langle (x^0_i)^2 \rangle_t = 1$, we have that $(\sigma^l)^2 = 2$.

\subsection{ Sample Mean and Variance }
In this paper, we will analyze the behavior of Kaiming initialization using the \emph{sample} mean and variance of pre-activations, averaged over network configurations:
\begin{subequations}
    \begin{align}
        \text{network averaged squared sample mean: } (m^l_i)^2 &= \langle \langle u^l_i \rangle_t^2 \rangle_w \\
        \text{network averaged sample variance: } (v^l_i)^2 &= \langle \langle (u^l_i)^2 \rangle_t - \langle u^l_i \rangle_t^2 \rangle_w
    \end{align}
\label{eqn:sample_stats}
\end{subequations}
As before, the average over $W$ implies these are the same for all $i$ within a layer so we can meaningfully define $m^l$ and $ v^l$ per-layer rather than per neuron. The critical difference between the two sets of statistics comes from the term $(m^l)^2$, the squared sample mean averaged over network configurations. Combining Equations \ref{eqn:total_stats} and \ref{eqn:sample_stats} and using the fact that $\mu^l=0$, we get the following relationship between the total and sample statistics:
\begin{equation}
    (m^l)^2 + (v^l)^2 = (\sigma^l)^2
    \label{eqn:total_sample}
\end{equation}
In other words, the total variance is the sum of two terms: the network-averaged sample variance and the network-averaged squared sample mean. Intuitively, randomness in a pre-activation's value has two sources: one from intrinsic randomness in the sample distribution and the other from randomness in the weight distribution. $(\sigma^l)^2$ combines the two sources while $(v^l)^2$ only gives the average size of fluctuations that arise only from randomness in the inputs. $(m^l)^2$ is the difference between the two.

\section{ Sample Variance Decays with Depth in ReLU MLPs }
\begin{figure}
\centering
\includegraphics[width=0.9\columnwidth]{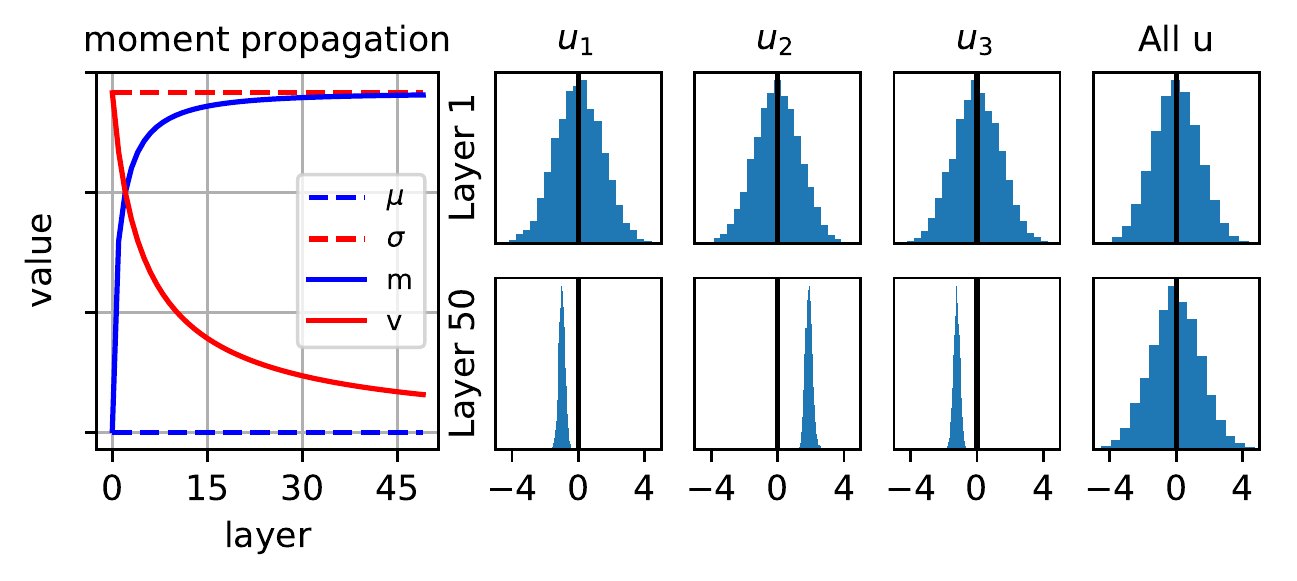}
\caption{ Left: Theoretical predictions for network-averaged sample mean ($m$) and standard deviation ($v$) vs. layer. Right: Pre-activation distribution in a wide ReLU MLP over samples (columns 1,2,3) and samples+networks (column 4). In layer 1 the distributions are similar. In layer 50 the sample variance has decayed significantly. }
\label{fig:theory_qualitative3}
\end{figure}
We first analytically predict the behavior of $m^l$ and $v^l$ in Kaiming-initialized MLPs in a restricted setting. We assume the sample distribution consists of input vectors whose elements are IID ($X^0_i \perp \!\!\! \perp X^0_j $ for $i \neq j$) and we assume that the MLPs are infinitely wide ($n^l \rightarrow \infty$). Just to simplify the arithmetic we additionally assume all layers are the same width ($n^l = n \rightarrow \infty$) and the inputs have been normalized so that $\langle X^0_i \rangle_t = 0 $ and $ \langle (X^0_i)^2 \rangle_t = 1$. In this simple setting, we show that the sample variance decays to zero with depth in Kaiming initialized networks.\footnote{One might expect this simple setting to be least likely to show non-trivial behavior.}

We then lift the restriction on infinite width and empirically determine the sample mean-standard deviation ratio in Kaiming initialized ReLU MLPs. We observe that the rate of sample variance decay is lessened by decreasing the network width and that increasing the width gives closer agreement between the infinite width calculation and experiment. Finally we lift both restrictions on infinite width and IID inputs and additionally allow more realistic architectures. We empirically determine the sample mean-standard deviation ratio in two commonly used architectures with two commonly used datasets and again find that sample variance decay is present in both networks.

\subsection{ Analytic Calculation: Wide Network, IID Inputs }

To calculate $m^l$ we first switch the order of expectation so that $ \langle \cdot \rangle_w $ is the inner expectation. Let $ u^l_{i,a} $ and $ u^l_{i,b} $ be the value of pre-activation $i$ in layer $ l $ for two independently chosen samples $\mathbf{x}^0_a$ and $ \mathbf{x}^0_b$. Similarly $ \mathbf{x}^l_a $ and $ \mathbf{x}^l_b $ are the values of the vector of activations in layer $ l $ for two independently chosen samples $\mathbf{x}^0_a$ and $ \mathbf{x}^0_b$. It can be shown that:
\begin{equation}
    (m^l)^2 = \langle \langle u^l \rangle_t^2 \rangle_w = \langle \langle u^l_a u^l_b \rangle_w \rangle_t = \frac{2}{n^{l-1}} \langle \langle \mathbf{x}^{l-1}_a \cdot \mathbf{x}^{l-1}_b \rangle_w \rangle_t
\end{equation}
We now apply the formalism of \citet{exponential_expressivity} to compute the inner product $\langle \mathbf{x}^{l-1}_a \cdot \mathbf{x}^{l-1}_b \rangle_w$ in a wide network for fixed $\mathbf{x}^0_a, \mathbf{x}^0_b$. Defining $c^l$ as the expected cosine similarity, we can apply their formalism to show:
\begin{equation}
    c^l := \langle \frac{\mathbf{x}^{l}_a \cdot \mathbf{x}^{l}_b} {|\mathbf{x}^{l}_a| \cdot |\mathbf{x}^{l}_b|} \rangle_w \;\;\;\;\;\;\;\;\;
    c^{l+1} = 2\int \mathcal{D} z_1 \mathcal{D}z_2 f(z_1) f(c^{l} z_1 + \sqrt{1-(c^{l})^2} z_2) := K(c^{l})
\label{eqn:K}
\end{equation}
where $ \mathcal{D}z_i = \frac{1}{2 \pi} e^{-z_i^2/2} $ are the unit Gaussian measures and $f$ is the ReLU activation. Given the cosine similarity between the input vectors $c^0$, we can compute $c^l$ at layer $l$ using the iterated map $K$: $c^l = K^l(c^0) = K \circ ... \circ K(c^0)$. The wide network assumption is critical in this step. In finite width networks, $\mathbf{x}^{l}_a \cdot \mathbf{x}^{l}_b$, $|\mathbf{x}^{l}_a|$, and $|\mathbf{x}^{l}_b|$ are (intractable) \emph{distributions} over network configurations, rather than deterministic functions of the inputs $\mathbf{x}^0_a$ and $ \mathbf{x}^0_b$. The insight from \citep{exponential_expressivity} was to realize these are self-averaging quantities, meaning that in the wide network limit, their distribution converges to a delta function around their average value. This insight allowed for a tractable analysis in wide feedforward networks, which we have used to write down Equation \ref{eqn:K}.

To get $ m^l $ we average this map $K$ over the distribution of sample cosine similarities: $ (m^l)^2 = 2\langle K^l(c^0) \rangle_t $. For arbitrary input distributions, this is a problematic integral. We again make use of the wide network assumption and we additionally use the IID input assumption to argue that the distribution of $\mathbf{x^0_a} \cdot \mathbf{x^0_b} $ becomes sharply peaked around 0 as $n^0 \rightarrow \infty $. This allows us to move the expectation inside the map: $ \langle K^l(c^0) \rangle_t \rightarrow K^l( \langle c^0 \rangle_t) $. 

Using the zero mean unit variance assumption on the input, we can see that $\langle c^0 \rangle_t = 0$ and $\sigma^2 = 2 \langle (x^0_i)^2 \rangle_t = 2$. This gives us the following equations for $m^l,v^l$ as a function of layer:
\begin{equation}
    (m^l)^2 = 2 K^l(0) \;\;\;\;\;\;\;\;\;\;\;\; (v^l)^2 = 2 (1 - K^l(0))
\end{equation}
We show the resulting dynamics of the network-averaged sample mean and variance alongside the total mean and variance in the left half of Figure \ref{fig:theory_qualitative3}. The distribution over samples for 3 randomly selected pre-activations and the distribution of a pre-activation over both samples and networks is displayed in the right half of Figure \ref{fig:theory_qualitative3}. One can see that in lower layers, a pre-activation's distribution over networks+samples and over just samples in a fixed network appear similar. As one looks at higher layers, the fluctuations of individual pre-activations in fixed networks over samples decays significantly while the fluctuations over samples and network configurations is well-preserved.

\subsubsection{ Slow Rate of Decay }
In Figure \ref{fig:theory_qualitative3} it appears that the sample standard deviation does not quite reach zero, even at depth 50. In this section we show that $v^l$ does in fact decay to zero as $L \rightarrow \infty $, but the rate of convergence is slow (subexponential). To show that $v^l \rightarrow 0$, we can use Equation \ref{eqn:K} to show that for all $ c \in [-1,1) $, $K(c) > c$. This implies that $c^{l+1} > c^l $. Additionally $K(1) = 1$. Therefore $c^l \rightarrow 1$ as $l \rightarrow \infty $ so that $v^l \rightarrow 0$ as $l \rightarrow \infty $. 

To show that the rate of convergence of $c^l$ to 1 is slow, we calculate the derivative of the iterated map, $K$ as $c$ approaches 1 from the left. We find that $ \lim_{c\rightarrow 1^-} dK(c)/dc = 1 $. This implies a subexponential convergence to the fixed point $c^*=1$. This is observed in the relatively slow decay of $v^l$ in Figure \ref{fig:theory_qualitative3}. It is interesting here that the decay is subexponential. For instance sending the variance of the initial weight distribution to $ \gamma \frac{2}{n} $ for $\gamma>1$, causes the total variance decays/explodes \emph{exponentially}.

\subsection{ Numerical Simulation: Finite Width, IID Inputs }
\begin{figure}
\centering
\includegraphics[width=0.9\columnwidth]{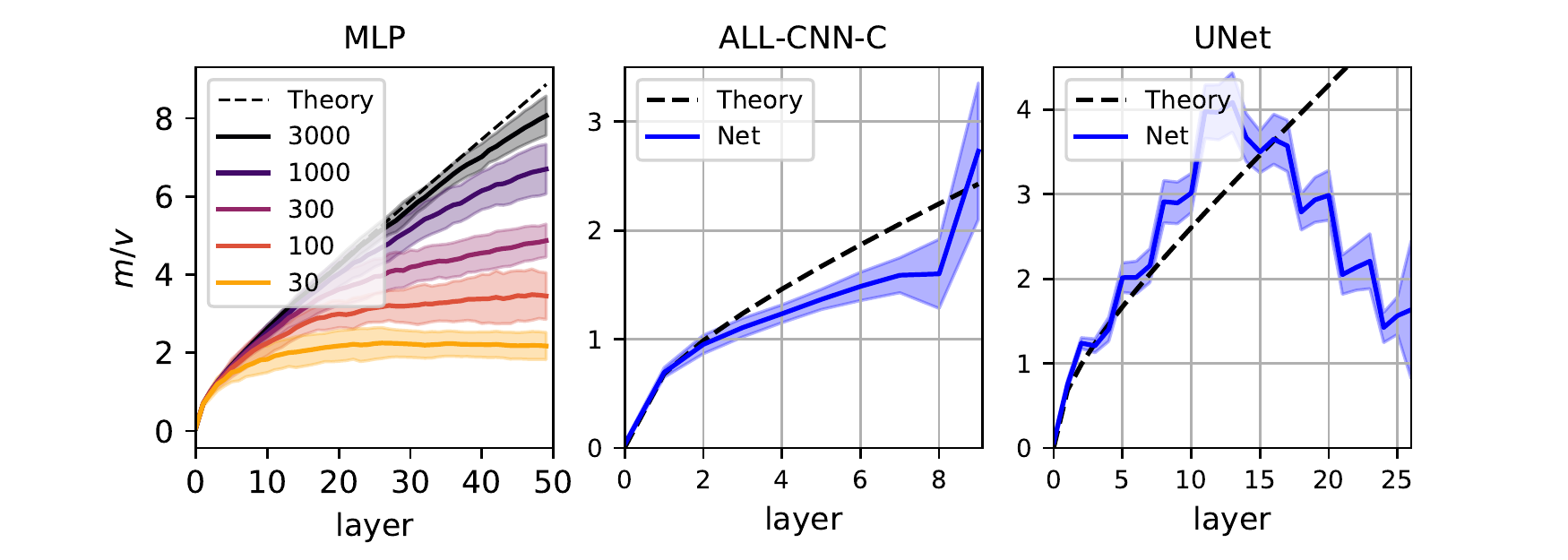}
\caption{ Sample mean-to-standard deviation ratio in finite networks. Solid line indicates average ratio over 30 random configurations and shaded represents standard deviation of ratio over configurations. }
\label{fig:finite_nets}
\end{figure}

We test the validity of the infinite width approximation for ReLU MLPs of finite width. We sample 30 random MLPs of depth 50 with uniform width, $ n $, per layer. We compare $ n = 30,100,300,1000,3000$. We sample 100 input vectors with elements chosen i.i.d. from an $ \mathcal{N}(0,1) $ distribution. These are then propagated through each network and the empirical ratio of squared sample mean to sample variance for each network is computed:
\begin{equation}
    r^l = \sqrt{\frac{\sum_{i=1}^{n^l} \langle u^l_i \rangle_t^2}{ \sum_{i=1}^{n^l} \langle (u^l_i)^2 \rangle_t - \langle u^l_i \rangle_t^2 }}
\end{equation}
The mean and standard deviation of this empirical ratio over the 30 randomly sampled networks is shown in the left part of Figure \ref{fig:finite_nets}. The dashed line shows the theoretical prediction for networks of finite width. It is observed to be an upper bound on the ratio observed in finite width networks. For a particular layer, as the network width increases, the empirical ratio approaches the theoretical ratio. 

\subsection{ Numerical Simulation: Real-World Convolutional Networks, Real-World Inputs }
Real world training scenarios of course use finite width networks and have samples with non-independent elements. Additionally, most architectures are more complicated than the uniform width MLPs we have so far described. Here we examine the behavior of the the empirical sample mean-std ratio in two commonly used networks on two datasets: ALL-CNN-C \citep{all_conv} on the CIFAR10 object recognition dataset and UNet \citep{unet} on the ISBI2012 neuron segmentation dataset \citep{crowdsourcing}.

The ALL-CNN-C network contains 9 layers of convolution, a global pooling layer, and a single fully connected output layer. We use reflection padding wherever necessary. The CIFAR10 dataset contains 32x32x3 images, each containing one of 10 objects. UNet contains 23 layers of convolutions and 4 layers of max pooling. Additionally it contains skip connections across layers. Instead of using "valid" convolutions, we use reflection padding on each feature map to preserve the spatial size after each convolution operation. The ISBI2012 dataset contains 512x512 microscopy images of neurons.

The middle and right parts of Figure \ref{fig:finite_nets} show the results for 30 random initializations. We see that the empirical ratio increases in all layers of ALL-CNN-C. The last layer actually has a higher empirical ratio than a wide MLP would. We attribute this to the global pooling which greatly increases the mean-std ratio. In UNet, the empirical ratio increases up to around layer 12 before decreasing significantly. We explain this decrease primarily by the skip connections from earlier to later layers. Additionally, we see that up to layer 12, the observed ratio is higher than predicted. We attribute this to the max pooling layers which increase the ratio more than a single ReLU layer.

\subsection{ Qualitative Explanation for Sample Variance Decay }
We provide an informal explanation for why the sample variance decays in ReLU networks, even when the total variance is preserved. The first observation is that the sample mean-std ratio of a set of vectors is on average preserved by matrix multiplication by a matrix with IID elements: 
\begin{equation}
    \frac{ \langle \langle W \mathbf{x} \rangle_{t} \cdot \langle W \mathbf{x} \rangle_{t} \rangle_w } {\langle \langle W \mathbf{x} \cdot W \mathbf{x} \rangle_{t} - \langle W \mathbf{x} \rangle_{t} \cdot \langle W \mathbf{x} \rangle_{t} \rangle_w} = \frac{ \langle \mathbf{x} \rangle_{t} \cdot \langle \mathbf{x} \rangle_{t}} {\langle \mathbf{x} \cdot \mathbf{x} \rangle_{t} - \langle \mathbf{x} \rangle_{t} \cdot \langle \mathbf{x} \rangle_{t}} \;\;\;\;\;\; \
     W_{ij} \stackrel{iid}\sim \mathcal{N}(0,1)
\end{equation}

The second observation is that applying the ReLU activation to any random variable (with negative support) \emph{increases} its mean and \emph{decreases} its variance:
\begin{equation}
    \int_{-\infty}^{\infty} P(x) f(x) dx > \int_{-\infty}^{\infty} P(x) x dx \;\;\; \text{ and } \ \int_{-\infty}^{\infty} P(x) f(x)^2 dx < \int_{-\infty}^{\infty} P(x) x^2 dx 
\end{equation}

where $ f $ is the ReLU function and $ P(x) > 0$ for some finite set of $x$

Because a feedforward ReLU network with zero-initialized biases is simply repeated application of matrix multiplication and ReLU nonlinearity, we should expect the magnitude of the mean over samples of individual pre-activations to increase, while the variance over samples decreases with layer. The precise \emph{rate} at which this occurs requires more careful analysis and was calculated earlier in this section.

\section{ Batch Normalization Eliminates Sample Variance Decay }
\begin{figure}
\centering
\includegraphics[width=0.85\columnwidth]{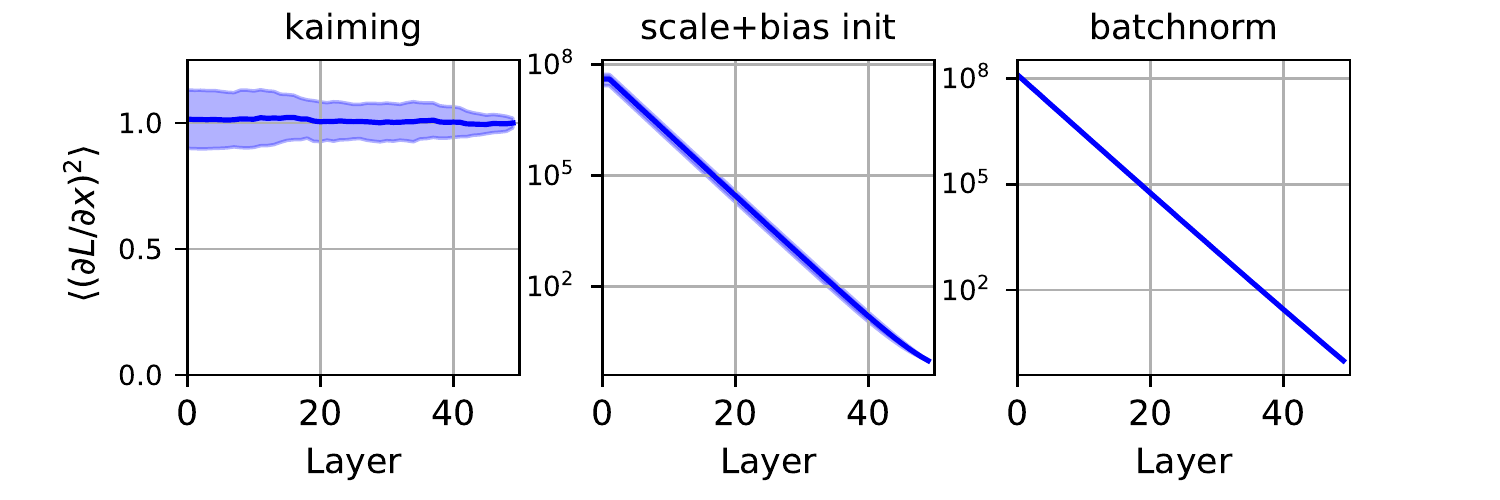}
\caption{ Gradients vs. layer. The scale of gradients remains fixed when the sample variance decays(Kaiming init, right). The gradients explode when the sample variance is preserved (scale+bias init, middle) and (batchnorm, right). }
\label{fig:gradients}
\end{figure}

Batch Normalization is a popular technique that is often found to improve training speed, allow for higher learning rates, reduce sensitivity to initialization, and improve generalization performance in a variety of settings \citep{batchnorm}. The precise reason for these benefits however is still unknown. Recent work has attempted to explain its success as a result of the \emph{reparameterization} \citep{batchnorm_optim}. 

We show here that Batch Normalization also qualitatively changes a deep ReLU network's \emph{initialization} by showing that, in the large batch limit at least, Batch Normalization eliminates sample variance decay at initialization. For simplicity we assume a large batch size so that sample statistics are well approximated by batch statistics. By construction, Batch Normalization subtracts the mean $\mu_s$ and divides by the standard deviation $\sigma_s$ for every pre-activation in the network:
\begin{equation}
    u^l_i \leftarrow (u^l_i- (\mu_s)^l_i) / (\sigma_s)^l_i
\end{equation}
Importantly these are the mean and standard deviation, over \emph{samples}, for a fixed network. Networks using Batch Normalization therefore have zero sample mean and fixed sample variance for all layers at initialization.

An informal argument using the calculation presented in Section 3 suggests that $\sigma_s = \sqrt{1-K(0)}=0.826$ in a wide ReLU network. In other words the effect of the rescaling should be to multiply the pre-activations at every layer by $1/0.826=1.21$ to counteract the sample variance decay from one layer of matrix mulplication+ReLU that was discussed in the previous section. A consequence of this scaling is that now in the backward pass, the gradients, which were originally preserved in magnitude with Kaiming initialization \citep{kaiming_init}, get amplified at every layer by $\sigma_s=1.21$. \emph{Rescaling to prevent sample variance decay causes gradients to explode exponentially}.

We support this informal argument with numerical simulations shown in Figure \ref{fig:gradients}. We sample 30 random MLPs of depth 50 and width 3000 and backpropagate using a random linear loss: $L=\mathbf{w} \cdot \mathbf{x^L} $ where $ w_{i} \stackrel{iid}\sim \mathcal{N}(0,1) $ and $\mathbf{x^L}$ is the output activation layer. The network and sample average of gradients $\langle (\partial L / \partial x^l)^2 \rangle_{wt} $ is plotted vs. layer $l$. We do this for 3 settings: Kaiming initialization, scale+bias initialization (described in Section 5), and Kaiming+Batch Normalization.

We estimate the slope of $ \log(\langle (\partial L / \partial x^l)^2 \rangle_{wt}) $ vs. $l$ shown in the middle (scale+bias) and right (Batch Normalization) graphs of Figure 3 as -0.379 and -0.381 respectively. The theoretical prediction using $\sigma_s=1.21$ is $-\log(\sigma_s^2) = -0.383 $. This informal analysis did not account for backpropagating through the sample variance $\sigma_s$ yet it still gave accurate prediction of the gradient explosion factor.

\section{ Training Impact }
We have shown the sample mean and variance of neurons in ReLU networks is not preserved with depth, even when total variance is. We have also shown that, in wide ReLU MLPs, Batch Normalization qualitatively changes the initialization by zeroing the sample mean, eliminating sample variance decay, and placing the network into a regime where the gradients grow exponentially. Is this just a theoretical curiosity? Do any of these observations impact network training? 

One the one hand it has long been argued that centering pre-activations should be helpful for training \citep{centering}. Possible benefits include improved conditioning of the loss surface and reduced noise in stochastic gradient updates. On the other hand, avoiding an extreme difference in the scale of gradients at various layers seems beneficial for training and the initialization scheme of \citet{xavier_init} was designed to mitigate this. It is further unclear the extent to which either of these phenomenon occur after a few training steps. Our analysis as presented only applies to \emph{random} networks.

We therefore investigate the question of training impact empirically using a data-dependent initialization scheme to set both the weights and biases so that the sample mean of every pre-activation is zero while the sample variance remains fixed with layer. We compare this to an initialization which sets the biases to zero and only scales the weights to ensure the total variance is preserved and to networks using Batch Normalization. 

\subsection{ Initialization Details }
Recall the original motivation for Kaiming initialization: scale the weights so the total variance remains constant with layer. In the simple ReLU networks we examined so far, this meant the weights should have a variance of $ 2/n $. A number of architectural elements arise in real world networks (skip connections and pooling in our two experiments) that make this scaling no longer hold. To implement the total variance preservation principle, we simply set the scale of the weights at each layer empirically so that the variance (over all pre-activations and samples over a fixed set of data points) is fixed in every layer. Note that the scale is per-layer, rather than per-feature. This idea was proposed before, notably in \citet{all_you_need_good_init}.

For \emph{scale} initialization, we sample the weights iid from a unit gaussian distribution and then rescale them \emph{per-layer} so the following pre-activations have unit sample variance. The biases are set to 0
\begin{equation}
    W^l_{ij} \leftarrow W^l_{ij} \frac{1}{ \sqrt{((\sigma_B)^l)^2 + \epsilon}} \;\;\;\;\;\;\; \
    ((\sigma_B)^l)^2 = \frac{1}{n^lT} \sum_{i=1}^{n^l} \sum_{t=1}^T (u^l_{ti})^2
\end{equation}
where the sum over T indices the sum over 5 minibatches of data. $ \epsilon = 1e-5 $. Note that the weights in layer $ l $ must be rescaled \emph{after} the weights in layer $ l-1 $.

For \emph{scale+bias} initialization, we first set the biases \emph{per-feature} to subtract out the sample mean. Then the weights are set using the scale initialization shown above.
\begin{equation}
    b^l_i \leftarrow - (\mu_B)^l_i \;\;\;\;\;\;\;\; (\mu_B)^l_i = \frac{1}{T} \sum_{t=1}^T u^l_{ti}
\end{equation}
where the sum over T indices the sum over 5 minibatches of data. Critically, this is done per-feature.

\subsection{ Training Details }
We compare the training times of scale initialization, scale+bias initialization, and batchnorm on two tasks: object recognition on the CIFAR10 dataset using the ALL-CNN-C architecture and image segmentation on the ISBI dataset using the UNet architecture. These datasets and networks were described in Section 3. We additionally augment the CIFAR10 dataset by zero padding each image, 4 pixels per side, and randomly cropping 32x32 sections. We augment the ISBI2012 dataset using 90 degree rotations, reflections, and elastic warps described in \citet{unet}.

We train each setting using both SGD with momentum 0.9 and the Adam optimizer with $\beta_1=0.9, \beta_2=0.999, \epsilon=1e-8$. For SGD, we search over the learning rates \{0.0003, 0.001, 0.003, 0.01\} and for Adam we search over the learning rates \{0.00003, 0.0001, 0.0003, 0.001\} and choose the fastest converging learning rate. We use 3 random seeds for each setting of parameters. 

\subsection{ Results }
\begin{figure}
\centering
\includegraphics[width=\columnwidth]{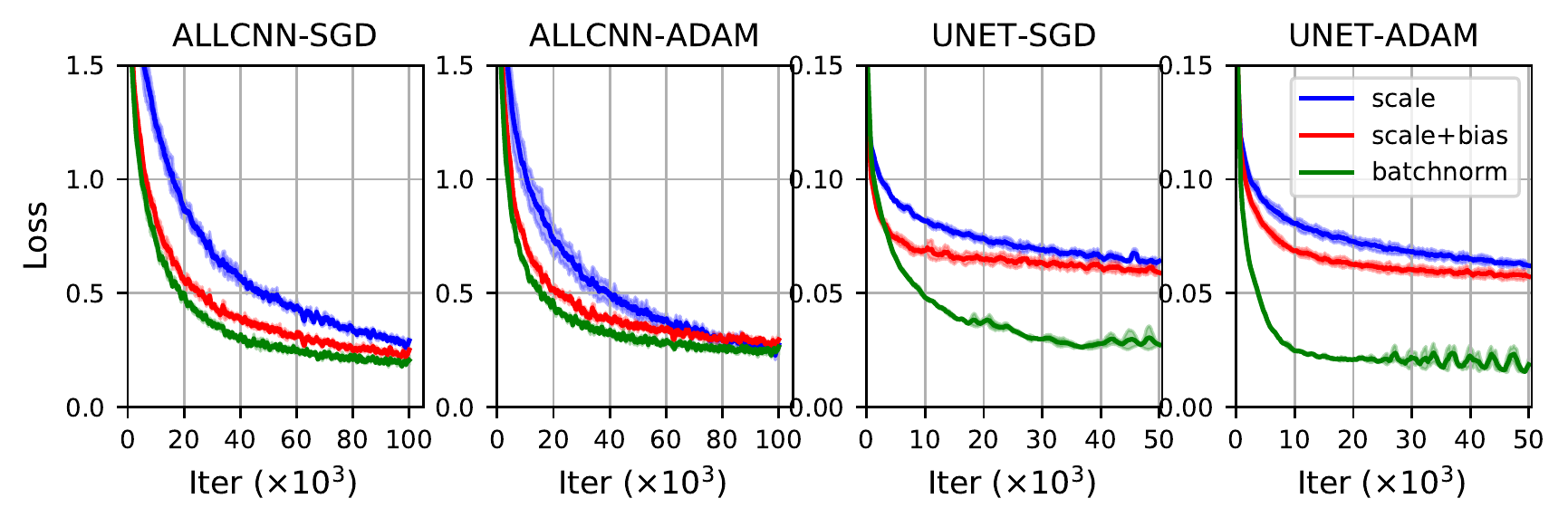}
\caption{ Training loss vs. iteration. The sample variance preserving "scale+bias" initialization scheme improved training speed over total variance preserving "scale" initialization scheme in each case. }
\label{fig:learning_curves}
\end{figure}
The results are shown in Figure \ref{fig:learning_curves}. The learning curves and their standard deviation over 3 randomly initialized runs are plotted. We observe in all cases, simply initializing the biases results in a gain in training speed.  It is interesting to compare the empirical ratio between the scale and bias+scale initialized network after training. For both tasks, the ratio in each setting converged to similar values in all layers after a few thousand iterations of sgd. We observe however that the loss for the bias+scale initialized networks is lower than the scale-only initialized networks well after 10k iterations in each case. It seems the network takes a long time to recover from the large ratio initialization.

\section{ Discussion }
We have shown that in wide Kaiming-initialized ReLU MLPs, all inputs are mapped to small fluctuations around some large input-independent output. This suggests that the network is operating in a "nearly linear" regime in higher layers. Higher layer pre-activations are either either always on or always off. This observation is consistent with the observation that that the scale of gradients are preserved with depth. By centering each pre-activation, either with Batch Normalization or the scale+bias initialization scheme, we ensure at initialization time that the nonlinearity of each pre-activation is used at every layer. It is in this regime that we observe exploding gradients: the network is implementing a highly nonlinear function.

These different regimes were actually studied by a number of papers including \citep{exponential_expressivity, deep_info_prop}. The preserved gradient regime can be identified with the "ordered" regime and the exploding gradient regime with the "chaotic" regime. Our analysis is closely related and in fact we used their formalism in our calculations.  By explicitly introducing sample randomness into our analysis, we have made the connection between between the formalism of \citep{exponential_expressivity} and the "total variance preservation" calculation of \cite{kaiming_init} more clear. It allowed us to make the non-trivial observations that the per-feature centering and per-layer scaling of Batch Normalization qualitatively change the initialization of a ReLU network. It allowed for a precise calculation of the impact of this re-initialization on the gradients of a wide ReLU MLP.



\bibliography{refs}
\bibliographystyle{plainnat}




\end{document}